# Blinded Radiologist and LLM-Based Evaluation of LLM-Generated Japanese Translations of Chest CT Reports: Comparative Study

Yosuke Yamagishi, MD, MSc [1]; Atsushi Takamatsu, MD, PhD [2, 3]; Yasunori Hamaguchi, MD [1];
Tomohiro Kikuchi, MD, PhD, MPH [4, 5]; Shouhei Hanaoka, MD, PhD [1, 2];
Takeharu Yoshikawa, MD, PhD [4]; Osamu Abe, MD, PhD [1, 2]

[1] Division of Radiology and Biomedical Engineering, Graduate School of Medicine, The University of Tokyo, Tokyo, Japan
[2] Department of Radiology, The University of Tokyo Hospital, Tokyo, Japan
[3] Department of Radiology, Kanazawa University Graduate School of Medical Sciences, Ishikawa, Japan
[4] Department of Computational Diagnostic Radiology and Preventive Medicine, The University of Tokyo Hospital, Tokyo, Japan
[5] Department of Radiology, School of Medicine, Jichi Medical University, Tochigi, Japan

## Abstract

**Background:** Accurate translation of radiology reports is important not only for multilingual research and clinical communication but also for radiology education and trainee learning. Recent large language models (LLMs) have demonstrated strong performance in medical translation, but the validity of LLM-based evaluation for educationally relevant translation quality remains unclear.

**Objective:** To evaluate the educational suitability of LLM-generated Japanese translations of chest CT reports using blinded radiologist assessment and to compare expert ratings with LLM-as-a-judge evaluations.

**Methods:** We used 150 chest CT reports from the validation set of CT-RATE-JPN. For each original English report, we compared a human-edited Japanese translation with an LLM-generated translation produced by DeepSeek-V3.2. A board-certified radiologist (radiologist 1) and a radiology resident (radiologist 2) independently performed blinded pairwise evaluations across four criteria: terminology accuracy, readability and fluency, overall report quality, and radiologist-style authenticity (which translation appeared more like a report written by a radiologist). In parallel, three LLM judges, DeepSeek-V3.2, Mistral Large 3, and GPT-5, evaluated the same translation pairs using identical criteria. Agreement between radiologists and between radiologists and LLM judges was assessed using quadratic weighted kappa (QWK) and percentage agreement.

**Results:** Inter-rater agreement between radiologists and LLM judges was near zero (QWK = -0.04 to 0.15), whereas some LLM-LLM pairs showed higher agreement (up to QWK = 0.29). Agreement between the two radiologists remained poor (QWK = 0.01 to 0.06). The radiologist 1 found the two translations nearly equivalent in terminology (59% tie) but favored the LLM translation for readability (51%) and overall quality (51%). The radiologist 2 found translations equivalent in readability (75% tie) but favored the human-edited translation for overall quality (40% vs 21%). All three LLM judges overwhelmingly and systematically favored the LLM translation across all criteria, with LLM preference rates ranging from 70% to 99% depending on the criterion, yielding near-zero agreement with either radiologist. For radiologist-style authenticity, all three LLM judges overwhelmingly rated the LLM translation as more radiologist-like (>93% of cases); the radiologist 2 more often rated the human-edited translation as more radiologist-like (66% of definitive responses; p = .002), whereas the radiologist 1 more often rated the LLM translation as more radiologist-like (64% of definitive responses; p = .002).
**Conclusions:** LLM-generated translations of chest CT reports were rated highly natural and fluent by both LLM judges and one human evaluator, but the two radiologists differed substantially in their assessments. LLM-as-a-judge evaluations showed a strong directional preference for the LLM translation and negligible agreement with radiologists. For educational deployment of translated radiology reports, automated LLM-based evaluation alone is insufficient; expert radiologist review remains important.

## Introduction

Accurate translation of medical documentation is a cornerstone of modern globalized healthcare, facilitating not only cross-border clinical communication but also medical education and patient empowerment [1–3]. In the era of data-driven medicine, the availability of high-quality translated datasets has become even more critical. Large-scale English-language medical datasets are increasingly being leveraged to develop AI models in other languages through machine translation, creating a pressing need for translation methods that maintain clinical nuance and technical precision [4–7].

Large Language Models (LLMs) have recently emerged as transformative tools in the field of medical natural language processing [8–10]. Recent studies suggest that state-of-the-art LLMs can achieve translation quality comparable to professional translators in general medical contexts, including patient-facing instructional materials and the simplification of discharge documentation [11,12]. However, the application of LLMs to highly specialized fields like radiology remains relatively limited [13]. Radiology reports contain dense technical terminology, substantial stylistic variation despite ongoing efforts toward standardization, and highly consequential clinical information for which even subtle wording differences may affect interpretation [14,15]. To our knowledge, research specifically evaluating the English-to-Japanese translation of chest CT reports remains scarce, warranting a more rigorous assessment of the latest models.

A significant bottleneck in advancing medical translation is the challenge of quality evaluation. Traditional automated metrics, such as Bilingual Evaluation Understudy (BLEU) [16] and Recall-Oriented Understudy for Gisting Evaluation (ROUGE) [17], rely primarily on lexical or n-gram overlap and often fail to capture semantic equivalence, factual consistency, or clinical accuracy [18]. In medical text evaluation, quality assessment is inherently challenging because semantically equivalent and clinically acceptable outputs may differ considerably in wording, making lexical-overlap-based evaluation inherently limited [19], and evaluating whether a translation preserves radiologist-aligned phrasing and reporting style may be even more difficult. Consequently, the "LLM-as-a-judge" framework, in which an LLM evaluates the output of another model, has gained traction as a scalable alternative to labor-intensive human review [20,21]. However, its validity in high-stakes radiology translation, particularly its alignment with the nuanced judgments of human specialists, remains insufficiently validated.
In this study, we evaluate the performance of DeepSeek-V3.2, a state-of-the-art open-weight LLM [22], in translating chest CT reports from English to Japanese. By comparing model-generated translations with human-edited versions from the CT-RATE-JPN dataset [23], we aim to assess the clinical utility of the latest LLMs for medical education and research. Furthermore, we critically examine whether the LLM-as-a-judge framework provides a reliable surrogate for expert evaluation, or whether it introduces systematic biases that diverge from the professional standards of a board-certified radiologist and resident.

## Methods

### Ethical Considerations

Given that this study utilized a publicly available dataset with deidentified patient information, and that our research focused on the translation and linguistic analysis of the existing dataset without accessing any additional patient data, institutional review board approval was not required for this research.

### Dataset

We used 150 chest CT reports from the validation set of CT-RATE-JPN [23], which is derived from CT-RATE [24], a publicly available dataset comprising chest CT imaging studies with corresponding free-text radiology reports. For each report, two Japanese translations were available:

- Human-edited translation: a translation produced through a multi-stage quality control pipeline. An initial machine translation was generated by GPT-4o mini, then reviewed and corrected by a radiology resident, and finally refined by a board-certified radiologist. This multi-stage process serves as the ground-truth Japanese translation in CT-RATE-JPN.
- LLM-generated translation: produced by DeepSeek-V3.2, a large language model with strong multilingual and medical domain performance.

Each evaluation instance consisted of the original English report, the human-edited Japanese translation, and the LLM-generated Japanese translation. The presentation order of the two Japanese translations (left/right or A/B) was randomized and counterbalanced.

### Translation Generation

LLM-generated translations were produced using DeepSeek-V3.2 [22]. DeepSeek-V3.2 was selected for its strong performance on multilingual medical tasks and its permissive redistribution license (Apache-2.0), which enables future dataset sharing. Translations were generated at temperature 0 using a standardized prompt instructing the model to produce an idiomatic Japanese radiology report from the English source. The full prompt is provided in Supplementary Material 1.

### Linguistic Analysis of Translation Pairs

To quantify text-level differences between the two translation sets, we performed a series of paired analyses on all 150 report pairs. Character count was computed after removing whitespace and newlines. Sentence segmentation was performed by splitting on Japanese sentence-ending punctuation and newlines, retaining segments of more than two characters. For morphological analysis, we used the Janome tokenizer [25] to tokenize each report and computed type-token ratio (TTR), the ratio of unique token types to total tokens, as a measure of lexical diversity, using both all parts of speech and content words only (nouns, verbs, adjectives, and adverbs). Lemma-level base forms were used where available. Statistical comparisons between paired human-edited and DeepSeek translations were performed using the Wilcoxon signed-rank test and paired t-test; two-tailed p-values are reported.

### Blinded Radiologist Evaluation

The translation pairs were independently evaluated by two radiologists: one board-certified radiologist (radiologist 1) and one radiology resident (postgraduate year 5, radiologist 2). Both evaluators were blinded to whether each translation was human-edited or LLM-generated.

#### *Evaluation Procedure*

For each of the 150 report pairs, evaluators were shown:

1. The original English chest CT report
2. Japanese Translation A
3. Japanese Translation B

The assignment of human-edited and LLM-generated translations to positions A and B was randomized. Evaluators selected their preferred translation or indicated equivalence for each criterion independently.

#### *Evaluation Criteria*

Evaluators assessed four criteria:

1. Terminology accuracy: accuracy of medical and anatomical terminology
2. Readability and fluency: naturalness and coherence of the translated text

3. Overall clinical suitability: overall appropriateness for use as a clinical radiology report
4. Radiologist-style authenticity: whether the translation more closely resembled the style and expression of a report written by a radiologist

For criteria 1–3, evaluators chose one of three options: Translation A is better, Translation B is better, or Equivalent. For criterion 4, evaluators chose Translation A, Translation B, or Equivalent, as this criterion was intended to capture stylistic impression rather than objective superiority. Responses were decoded post hoc to determine whether the human-edited or LLM-generated translation was preferred.

### LLM-as-a-Judge Evaluation

Three LLM judges evaluated the same 150 translation pairs using an identical prompt structure:

- DeepSeek-V3.2
- Mistral Large 3
- GPT-5

These models were selected to represent different evaluation perspectives: DeepSeek-V3.2 was included as a self-judge because it was also used as the translation model, GPT-5 was included as a high-end commercial model [26], and Mistral Large 3 was included as a comparatively neutral open-weight model [27]. Each judge received the original English report and the two Japanese translations in randomized order, with the translation source hidden. The judge was instructed to evaluate the same four criteria and provide a structured JSON response indicating which translation was preferred for each criterion (A / B / TIE), along with a brief justification. Temperature was set to 0 for all judge evaluations. The full judge prompt is provided in the Supplementary Material.

### Statistical Analysis

For each criterion and each evaluator, we report:

- Count and percentage favoring the human-edited translation
- Count and percentage favoring the LLM translation
- Count and percentage of equivalent (TIE) responses

Inter-rater agreement between the two radiologists, and between each radiologist and each LLM judge, was quantified using quadratic weighted kappa (QWK). QWK was chosen because the three response categories (human-edited preferred / equivalent / LLM preferred) form a natural ordinal scale; QWK appropriately penalizes larger disagreements more than adjacent-category disagreements. The ordinal order used was: human-edited preferred < equivalent < LLM preferred. Raw percentage agreement is also reported. For criterion 4 (radiologist-style authenticity), we treated definitive responses (excluding Uncertain) as binary outcomes and tested whether preference for the human-edited translation exceeded chance (50%) using a one-sided binomial test. This tests whether the human-edited translation was more consistently judged as radiologist-like, not whether evaluators could correctly identify the translation method.

## Results

### Linguistic Comparison of Translation Pairs

Human-edited translations were significantly longer than DeepSeek-generated translations in total character count (median 517.5 vs. 481.5 characters; Wilcoxon signed-rank test, $W = 729.5$, $p < .001$), with DeepSeek translations averaging approximately 94% of the length of their human-edited counterparts (mean ratio 0.940). Despite this overall brevity, DeepSeek translations contained significantly more sentences per report (mean 20.3 vs. 19.6; $p < .001$), while individual sentences were significantly shorter (mean sentence length 24.7 vs. 27.3 characters; $p < .001$). This pattern indicates that DeepSeek-generated translations segmented content into a greater number of shorter sentences compared with the human-edited versions. Lexical diversity, measured by TTR, differed significantly between the two translation sets when computed across all parts of speech: DeepSeek translations showed higher TTR (mean 0.402 vs. 0.381; $p < .001$), indicating greater lexical variety relative to total token count. This seemingly counterintuitive finding is partly attributable to the shorter overall length of DeepSeek translations, as TTR tends to increase with shorter texts. When TTR was restricted to content words only (nouns, verbs, adjectives, and adverbs), no significant difference was observed (mean 0.550 vs. 0.549; $p = .665$), suggesting that the two translation methods employed a comparable breadth of clinically meaningful vocabulary.

### Radiologist Evaluation Results

Figure 1 summarizes the blinded radiologist evaluation results.

#### *Terminology Accuracy*

The radiologist 1 rated 59% (n=89) of report pairs as equivalent in terminology, favoring the LLM translation in 23% (n=35) and the human-edited translation in 17% (n=26) of cases. The radiologist 2 rated 51% (n=76) as equivalent, favoring the human-edited translation in 35% (n=53) and the LLM translation in 14% (n=21) of cases.

#### *Readability and Fluency*

The radiologist 1 favored the LLM translation for readability in 51% (n=77) of cases, with human-edited preferred in 25% (n=38) and equivalent in 23% (n=35). The radiologist 2 found translations equivalent in readability in 75% (n=112) of cases, favoring LLM in 15% (n=22) and human-edited in 11% (n=16).

### *Overall Report Quality*

The radiologist 1 favored the LLM translation for overall quality in 51% (n=77) of cases, with human-edited preferred in 29% (n=43) and equivalent in 20% (n=30). The radiologist 2 favored the human-edited translation in 40% (n=60) of cases, found translations equivalent in 39% (n=58), and favored the LLM translation in 21% (n=32).

### *Radiologist-Style Authenticity*

When asked which translation more closely resembled the style of a report written by a radiologist, the radiologist 2 more often rated the human-edited translation as more radiologist-like (66% of definitive responses; binomial test $p = .002$). In contrast, the radiologist 1 more frequently rated the LLM translation as more radiologist-like (64% of definitive responses; $p = .002$).

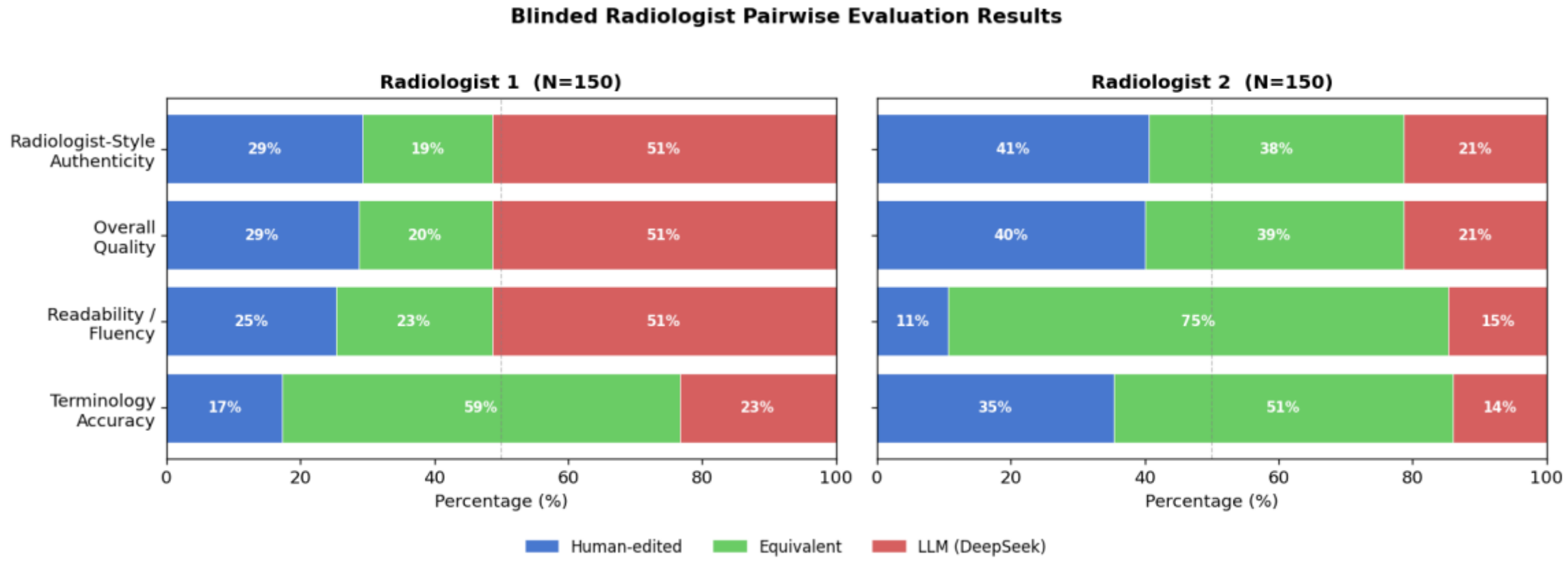

Figure 1. Blinded radiologist pairwise evaluation results. Stacked bar charts for radiologist 1 and radiologist 2 across four criteria.

### *Inter-Rater Agreement*

Inter-rater agreement between the two radiologists was poor across all four criteria (confusion matrices in Figure 2). QWK ranged from 0.012 (radiologist-style authenticity) to 0.059 (readability/fluency), and raw agreement ranged from 28% (readability) to 37% (terminology accuracy). These values are consistent with slight-to-no agreement, well below the threshold conventionally considered as fair agreement (QWK ≥ 0.20).

The direction of disagreement was notably systematic: the radiologist 1 consistently favored the LLM translation more than the radiologist 2 did, particularly for readability and overall quality. For overall quality, the radiologist 1 favored the LLM in 51% of cases versus 21% for the radiologist 2; the radiologist 2 favored the human-edited in 40% versus 29% for the radiologist 1.

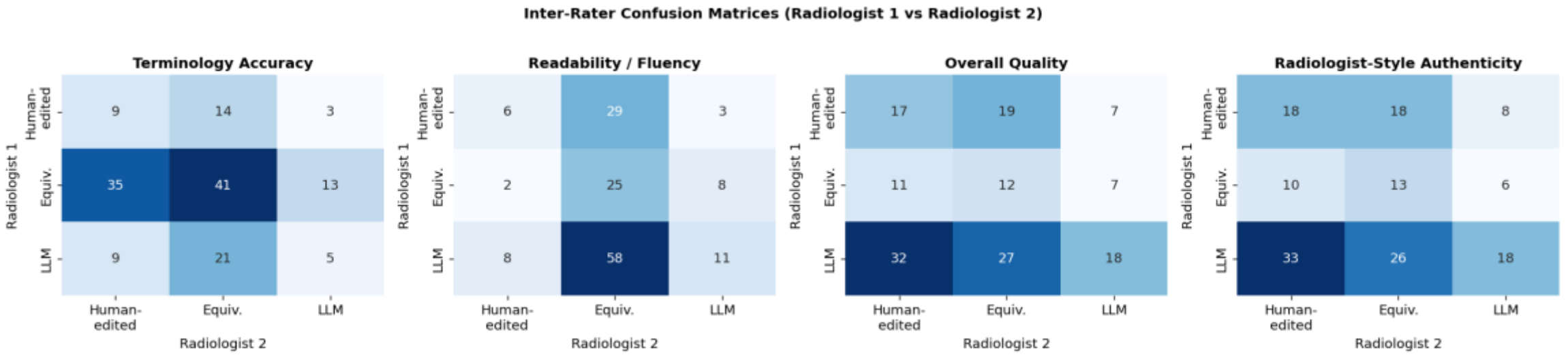

Figure 2. Inter-rater confusion matrices between radiologist 1 and radiologist 2.

## LLM-as-a-judge Results

All three LLM judges strongly and systematically favored the LLM-generated translation across all criteria (Figure 3).

### *Terminology Accuracy*

Judges favored the LLM translation in 79–91% of cases (DeepSeek: 91%, n=137; Mistral: 79%, n=119; GPT-5: 81%, n=122). TIE rates were low (7–13%), and the human-edited translation was preferred in only 2–18% of cases.

### *Readability and Fluency*

Preference for the LLM translation was nearly universal: 70% (DeepSeek, n=105), 88% (Mistral, n=132), and 95% (GPT-5, n=142). Equivalent ratings were near zero.

### *Overall Report Quality*

Judges favored the LLM translation in 83–95% of cases (DeepSeek: 95%, n=142; Mistral: 91%, n=137; GPT-5: 83%, n=125). Human-edited was preferred in 5–17%.

### *Radiologist-Style Authenticity*

All three judges almost universally rated the LLM translation as more radiologist-like: DeepSeek favored LLM in 99% (n=148), Mistral in 93% (n=140), and GPT-5 in 95% (n=143) of cases. The human-edited translation was rated as more radiologist-like in <7% of responses for all three judges.

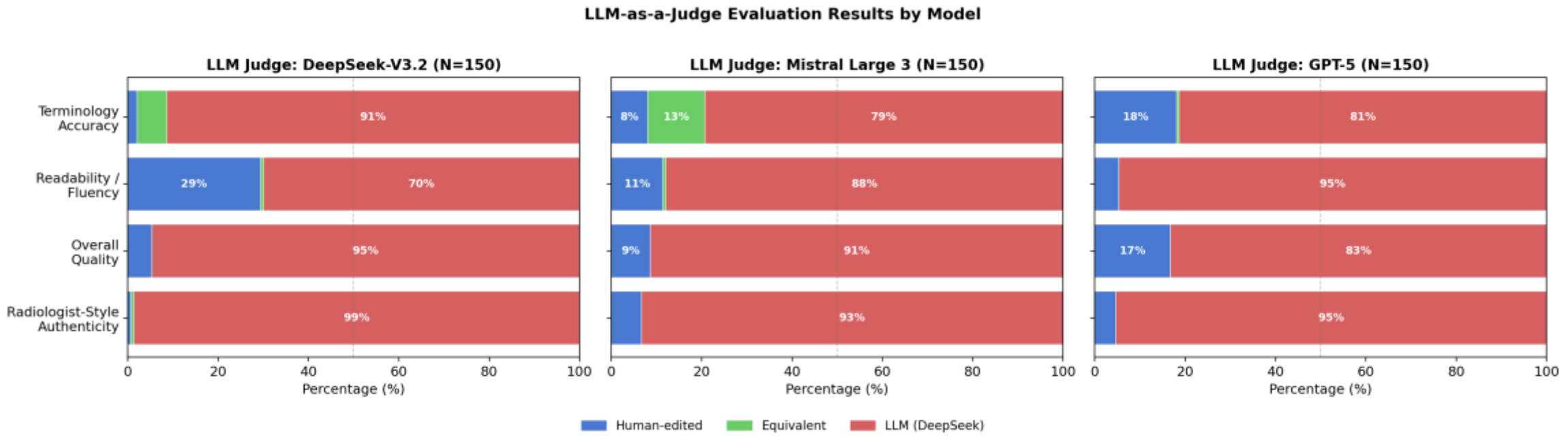

Figure 3. LLM-as-a-judge evaluation results for all three judge models.

## Radiologist vs LLM Judge Agreement

Agreement between radiologists and LLM judges was near zero across all criteria and criterion–evaluator combinations (Table 1, Figure 4). QWK between either radiologist and any LLM judge ranged from −0.038 to 0.148. Agreement among the LLM judges themselves ranged from poor to moderate depending on the criterion pair, with QWK values ranging from −0.010 to 0.286, but these judge–judge agreements did not translate into meaningful agreement with human radiologists.

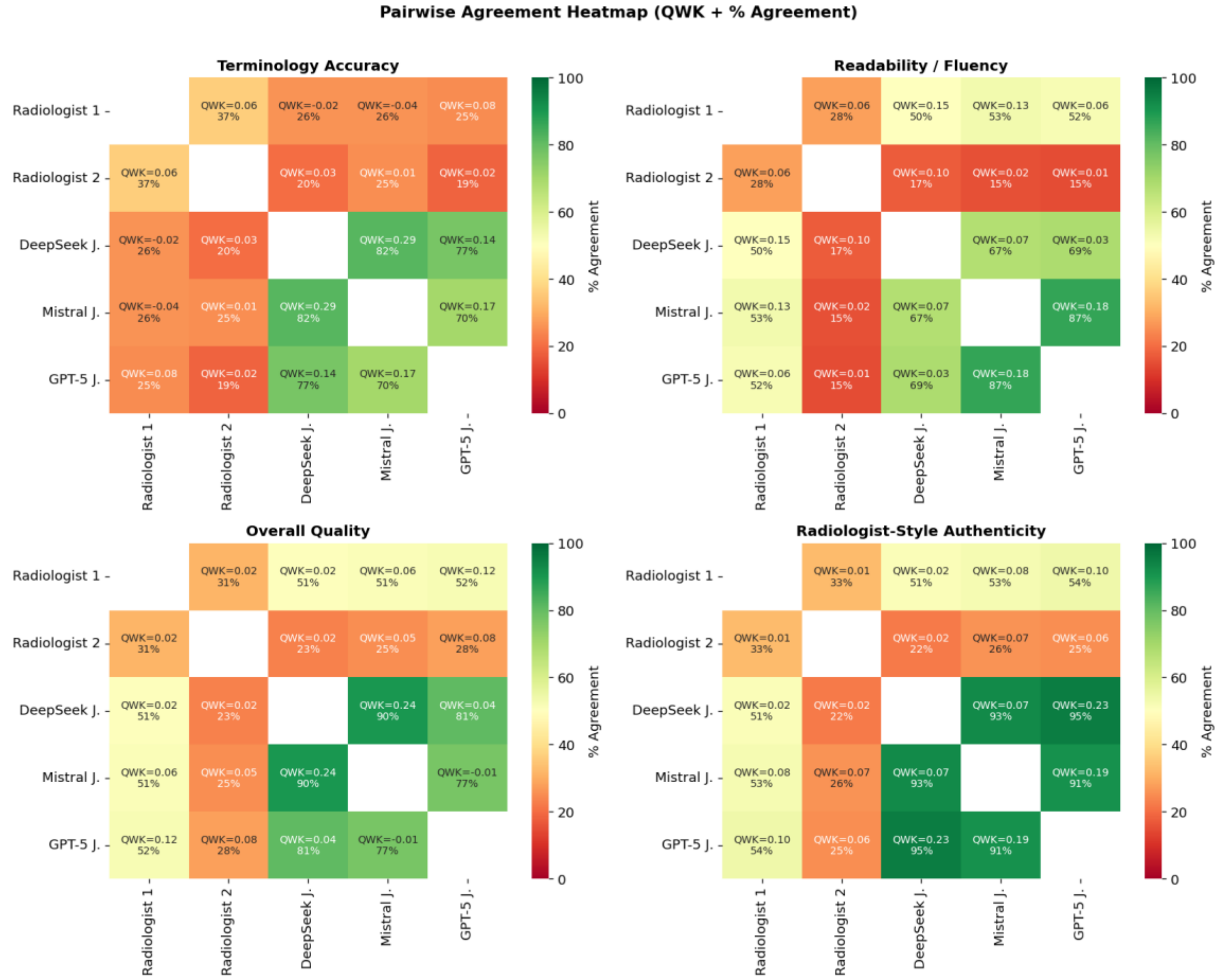

Figure 4. Pairwise agreement heatmap (QWK and percent agreement) for all evaluator pairs.

Table 1. Pairwise inter-rater agreement across radiologists and LLM judges for each evaluation criterion, expressed as QWK and percent agreement.

| Criterion | Rater 1 | Rater 2 | QWK | % Agreement |
|---|---|---|---|---|
| Terminology Accuracy | Radiologist 1 | Radiologist 2 | 0.056 | 36.7 |
| | Radiologist 1 | DeepSeek V3.2 | -0.022 | 26 |
| | Radiologist 1 | Mistral Large 3 | -0.038 | 26 |
| | Radiologist 1 | GPT-5 | 0.084 | 24.7 |
| | Radiologist 2 | DeepSeek V3.2 | 0.026 | 20 |
| | Radiologist 2 | Mistral Large 3 | 0.015 | 25.3 |
| | Radiologist 2 | GPT-5 | 0.025 | 18.7 |
| | DeepSeek V3.2 | Mistral Large 3 | 0.286 | 82 |
| | DeepSeek V3.2 | GPT-5 | 0.136 | 77.3 |
| | Mistral Large 3 | GPT-5 | 0.169 | 70 |
| Readability / Fluency | Radiologist 1 | Radiologist 2 | 0.059 | 28 |
| | Radiologist 1 | DeepSeek V3.2 | 0.148 | 50 |
| | Radiologist 1 | Mistral Large 3 | 0.128 | 53.3 |
| | Radiologist 1 | GPT-5 | 0.063 | 52 |
| | Radiologist 2 | DeepSeek V3.2 | 0.105 | 17.3 |
| | Radiologist 2 | Mistral Large 3 | 0.016 | 15.3 |
| | Radiologist 2 | GPT-5 | 0.007 | 14.7 |
| | DeepSeek V3.2 | Mistral Large 3 | 0.071 | 67.3 |
| | DeepSeek V3.2 | GPT-5 | 0.026 | 68.7 |
| | Mistral Large 3 | GPT-5 | 0.177 | 86.7 |
| Overall Quality | Radiologist 1 | Radiologist 2 | 0.021 | 31.3 |
| | Radiologist 1 | DeepSeek V3.2 | 0.016 | 50.7 |
| | Radiologist 1 | Mistral Large 3 | 0.055 | 51.3 |
| | Radiologist 1 | GPT-5 | 0.119 | 52 |
| | Radiologist 2 | DeepSeek V3.2 | 0.021 | 23.3 |
| | Radiologist 2 | Mistral Large 3 | 0.05 | 25.3 |
| | Radiologist 2 | GPT-5 | 0.076 | 28 |
| | DeepSeek V3.2 | Mistral Large 3 | 0.235 | 90 |
| | DeepSeek V3.2 | GPT-5 | 0.044 | 80.7 |
| | Mistral Large 3 | GPT-5 | -0.01 | 77.3 |
| Radiologist-Style Authenticity | Radiologist 1 | Radiologist 2 | 0.012 | 32.7 |
| | Radiologist 1 | DeepSeek V3.2 | 0.016 | 51.3 |
| | Radiologist 1 | Mistral Large 3 | 0.079 | 52.7 |
| | Radiologist 1 | GPT-5 | 0.105 | 54 |
| | Radiologist 2 | DeepSeek V3.2 | 0.016 | 22 |
| | Radiologist 2 | Mistral Large 3 | 0.069 | 26 |
| | Radiologist 2 | GPT-5 | 0.063 | 25.3 |
| | DeepSeek V3.2 | Mistral Large 3 | 0.072 | 92.7 |
| | DeepSeek V3.2 | GPT-5 | 0.229 | 95.3 |
| | Mistral Large 3 | GPT-5 | 0.191 | 91.3 |

## Qualitative Analysis of Radiologist and LLM Judge Disagreements

### *Representative Disagreement Cases*

To illustrate the nature of disagreements between radiologist evaluators and LLM judges, we identified cases in which both radiologists preferred the human-edited translation while all three LLM judges unanimously preferred the LLM-generated translation for the same criterion. The number of such cases was 5 (3.3%) for terminology accuracy, 4 (2.7%) for readability and fluency, 10 (6.7%) for overall quality, and 12 (8.0%) for radiologist-style authenticity; no case met this criterion across all four domains simultaneously.

Representative examples are instructive, and are drawn from cases flagged for terminology accuracy. In one case, the LLM-generated translation rendered "sequela fibrotic changes" as a term approximating "prognostic symptom," a clinically meaningful mistranslation. Despite this error, GPT-5 did not acknowledge it in its evaluation rationale, while DeepSeek-V3.2 and Mistral Large 3 explicitly characterized the terminology as accurate and praised it positively. In a second example, the Japanese term for the right lobe of the thyroid gland was rendered with reversed word order, deviating from standard Japanese anatomical nomenclature, yet all three LLM judges overlooked this error entirely. A third case involved the translation of "tree-in-bud appearance," a standard radiological term, as a literal Japanese rendering approximating "budding tree-like appearance." This phrasing is not used in clinical radiology reports in Japan, where the term is universally retained in English. Despite this, one LLM judge rated this translation favorably, while the remaining two did not explicitly penalize it. Both radiologists rated the human-edited translation as superior for this case.

### *Patterns in LLM Judge Rationales*

A review of the judges' free-text rationales in these disagreement cases suggested that stylistic preferences played a substantial role in LLM-based evaluations (Table 2, 3). Across cases where radiologists preferred the human-edited translation but LLM judges preferred the LLM-generated version, the terms *concise* and *natural* appeared repeatedly in the judge explanations. This pattern was especially prominent for readability/fluency and radiologist-style authenticity, whereas these expressions were rarely invoked for terminology accuracy and overall quality. As shown in Table 2, 3, *concise* was frequently cited when judges favored LLM-generated translations for readability/fluency and radiologist-style authenticity, particularly by GPT-5 and Mistral Large 3. A similar tendency was observed for *natural*, which was again concentrated in readability/fluency judgments, especially for GPT-5 and Mistral Large 3.

Table 2. Frequency of the term "concise" in LLM judges' rationales for disagreement cases.

| Judge | DeepSeek-V3.2 | | | Mistral Large 3 | | | GPT-5 | | |
|---|---|---|---|---|---|---|---|---|---|
| Winner | Human-edited | TIE | LLM | Human-edited | TIE | LLM | Human-edited | TIE | LLM |
| Terminology Accuracy | 0 | 0 | 0 | 0 | 0 | 0 | 0 | 0 | 0 |
| Readability / Fluency | 1 | 0 | 20 | 0 | 0 | 84 | 2 | 0 | 87 |
| Overall Quality | 0 | 0 | 3 | 0 | 0 | 16 | 0 | 0 | 0 |
| Radiologist-Style Authenticity | 0 | 0 | 70 | 4 | 0 | 119 | 1 | 0 | 23 |

Table 3. Frequency of the term "natural" in LLM judges' rationales for disagreement cases.

| Judge | DeepSeek-V3.2 | | | Mistral Large 3 | | | GPT-5 | | |
|---|---|---|---|---|---|---|---|---|---|
| Winner | Human-edited | TIE | LLM | Human-edited | TIE | LLM | Human-edited | TIE | LLM |
| Terminology Accuracy | 0 | 0 | 0 | 0 | 0 | 21 | 0 | 1 | 10 |
| Readability / Fluency | 17 | 0 | 42 | 11 | 0 | 55 | 8 | 0 | 105 |
| Overall Quality | 0 | 0 | 10 | 0 | 0 | 4 | 0 | 0 | 2 |
| Radiologist-Style Authenticity | 0 | 0 | 4 | 1 | 0 | 25 | 1 | 0 | 5 |

## Discussion

### Principal Results

This study found that LLM-generated Japanese translations of chest CT reports received divergent assessments depending on the evaluator. The radiologist 1 rated the LLM translation favorably for readability and overall quality, more so than the radiologist 2. The radiologist 2, by contrast, found translations frequently equivalent and showed a slight preference for the human-edited version overall. Crucially, all three LLM judges showed strong and systematic preference for the LLM translation, far exceeding even the more favorable of the two human assessments, and showed negligible agreement with either human rater.

The inter-rater agreement between the two radiologists was strikingly low (QWK = 0.01–0.06 for all criteria), underscoring that even expert radiologists do not apply identical standards when evaluating translation quality. Importantly, this low agreement should be interpreted in context rather than viewed simply as annotation failure. In post hoc debriefing, both radiologists reported that, in many cases, the two translations were nearly indistinguishable in overall quality, and that final preferences were often determined by subtle wording differences or minor nuances. Under such conditions, some degree of variability in pairwise preference judgments is expected. Nevertheless, this level of variability has important methodological implications: studies relying on a single human rater to validate LLM translation quality may produce conclusions that are specific to that rater's preferences rather than reflecting a shared clinical standard.

### LLM Translation Quality and Its Implications for Education and AI Development

The linguistic analysis revealed that LLM-generated translations were systematically shorter, contained more sentences, and had shorter individual sentences compared with human-edited counterparts. Despite this, content-word lexical diversity was comparable between the two methods, suggesting that the core clinical vocabulary was equivalently represented. These structural differences likely underlie the divergent stylistic impressions reported by the two radiologists.

From an educational standpoint, this study has direct relevance to the broader use of translated medical materials in training and cross-lingual learning contexts [1–3]. For educational contexts requiring precise clinical terminology and faithful representation of diagnostic uncertainty, an essential feature of radiology reporting [28,29], conciseness and natural phrasing, although desirable, should not be assumed to be the most important indicators of translation quality in medical texts.

Beyond medical education, the quality of machine-translated radiology reports has important implications for AI model development. Large-scale multilingual datasets are increasingly used to train and evaluate AI models for non-English clinical environments [4–7]. The fidelity with which LLM translation preserves clinical nuance, terminology conventions, and uncertainty expressions directly determines the quality of the training signal these datasets provide. Our findings suggest that LLM-generated translations are broadly adequate for high-volume dataset enrichment, such as pre-training or language model fine-tuning, where breadth and fluency are prioritized. However, for datasets intended to support high-stakes downstream tasks such as automated radiology report generation [30–32], expert review of at least a representative subset remains necessary to ensure that subtle clinical conventions are faithfully preserved.

### Why LLM Judges Diverge from Radiologists

By contrast, the LLM-based judges showed a much stronger and more directional preference pattern, favoring the LLM translation in 79–95% of cases across all criteria. This discrepancy suggests that the LLM judges may not have been detecting clinically meaningful superiority alone, but may also have reflected model-specific stylistic preferences or alignment with particular lexical and syntactic patterns, consistent with prior reports of self-preference bias in LLM-as-a-judge settings [33]. In other words, when candidate translations are broadly similar in quality, LLM-as-a-judge may amplify small textual differences into disproportionately confident preferences.

Several mechanisms may underlie this systematic bias. Prior work has shown that LLM-as-a-judge systems are susceptible to multiple forms of bias [34], raising the possibility that our three LLM judges were influenced by similar evaluation tendencies. As a result, they may have rewarded text that appeared polished, smooth, and lexically precise, even when those characteristics did not necessarily correspond to greater clinical appropriateness. A related explanation is that fluency may be implicitly conflated with overall translation quality [35]. Because contemporary LLMs are optimized in part through human preference signals that often reward naturalness and readability [36], they may systematically assign higher scores to outputs that sound more fluent on the surface, creating a structural advantage for LLM-generated translations [37].

Another likely factor is limited sensitivity to clinical register. Japanese radiology reports follow highly conventionalized patterns in wording, uncertainty marking, anatomic description, and overall sentence style. Even when a translation is grammatically correct and semantically plausible, small deviations from these conventions may render it less acceptable to radiologists. LLM judges may not be sufficiently grounded in radiology-specific reporting norms to recognize and appropriately weight such distinctions.

This interpretation is further supported by the marked discrepancy in judgments of radiologist-like style. All three LLM judges almost uniformly rated the LLM translation as more characteristic of radiologist writing (>93% of cases). The two radiologists, however, not only disagreed with the LLM judges but also with each other: radiologist 2 showed an above-chance preference for the human-edited translation ($p = .002$), while radiologist 1 showed the opposite preference ($p = .002$). This pattern suggests that the criteria used to assess radiologist-style authenticity vary substantially across evaluators—whether human or LLM—and that surface fluency alone may not capture what expert radiologists consider professionally authentic writing.

## Practical Recommendation

These findings point toward a tiered approach for deploying LLM-translated radiology reports. For applications where breadth and readability are prioritized, such as parallel corpora for language learning, multilingual teaching collections, or large-scale pre-training datasets for AI development, LLM-generated translations may be used with confidence. For higher-stakes applications, including model reports, teaching files, standardized assessment materials, or training data for clinical AI systems in which faithful representation of uncertainty is critical, expert radiologist review remains an important quality gate.

Critically, these findings indicate that LLM-as-a-judge should not serve as the sole quality gate for either educational content or AI training data. Given the near-zero agreement with radiologist assessments and the systematic tendency to favor LLM-generated output, automated judging appears to substantially overestimate educational and clinical suitability. A more practical approach is therefore a hybrid workflow: LLM translation and automated evaluation for initial scaling and screening, followed by targeted expert review of cases intended for high-stakes applications. This workflow balances scalability with the clinical rigor that radiology-specific use cases demand.

## Limitations

Several limitations should be noted. First, our dataset comprised only chest CT reports, and findings may not generalize to other modalities or report styles. Second, only Japanese translations were evaluated; results may differ for other target languages. Third, we had only two human raters, limiting statistical power for inter-rater analyses and precluding adjudication of disagreements. Fourth, the LLM-as-a-judge analysis was based on a single prompt framework and did not include systematic exploration of alternative judge prompts, rubric structures, or prompt optimization strategies. Fifth, we did not assess downstream educational outcomes—whether trainees who learn from LLM-translated versus human-edited reports show differences in knowledge acquisition or terminology proficiency. Such outcomes studies would be necessary to establish the ultimate educational utility of LLM-translated reports.

## Future Directions

Future work should investigate: (1) the impact of translation quality on trainee learning outcomes in controlled educational settings; (2) systematic exploration and optimization of LLM-as-a-judge prompts and rubric designs grounded in radiology terminology standards; (3) extension to other languages, modalities, and report types; and (4) the development of radiologist-aligned automated evaluation metrics for medical translation quality.

## Conclusions

LLM-generated Japanese translations of chest CT reports received divergent assessments from the two radiologists, with poor inter-rater agreement across all criteria. However, both human evaluators differed markedly from the three LLM judges, who showed systematic and extreme preference for the LLM translation far beyond either radiologist's assessment. LLM-as-a-judge evaluations showed near-zero QWK agreement with radiologist judgments and rated the LLM-generated translation as more radiologist-like in >93% of cases. These findings indicate that LLM-generated radiology report translations may hold promise for educational support, but that LLM-as-a-judge evaluation is insufficient for quality assurance in this domain. Expert radiologist review remains an important component of quality control when translating radiology reports for educational use.

## Data Availability

The dataset used in the present analyses was CT-RATE-JPN [38], a derived dataset based on CT-RATE [39]. CT-RATE-JPN is publicly available on Hugging Face. The original CT-RATE dataset is also publicly available on Hugging Face.

## Acknowledgements

The authors used GPT (OpenAI) and Claude (Anthropic) to assist with English-language editing and manuscript refinement. The authors reviewed and revised all AI-assisted outputs and take full responsibility for the final content of the manuscript.

## Conflicts of Interest

The Department of Computational Diagnostic Radiology and Preventive Medicine of The University of Tokyo Hospital is sponsored by HIMEDIC Inc and Siemens Healthcare KK.

## Author Contributions

YY contributed to the conceptualization, study design, implementation, formal analysis, and manuscript writing. AT and YH served as radiologist evaluators. TK, SH, TY, and OA provided supervision and critically reviewed the manuscript for important intellectual content. All authors reviewed, approved the final manuscript, and agree to be accountable for all aspects of the work.

## Abbreviations

BLEU: Bilingual Evaluation Understudy
CT: computed tomography
LLM: large language model
QWK: quadratic weighted kappa
ROUGE: Recall-Oriented Understudy for Gisting Evaluation
TTR: type-token ratio

## Supplementary Material

**Translation Prompt for DeepSeek-V3.2**

Translations were generated using the following system prompt at temperature 0. The model received the English report text as the user message, with no additional instructions.

**System prompt (Japanese original)**

```
あなたは放射線科の読影レポートを翻訳する専門家です。
以下の英語の読影レポートを、日本語として自然で臨床的に正確な表現に翻訳してください。

制約・注意:
- 医学的意味を変えない
- 放射線科で一般的な日本語表現を用いる
- コロンやセミコロンは使用せずに、「。」や「、」を使うようにする
- 「断層内」という表現を用いず、「撮像範囲内」や「検査範囲内」とする
- 造影剤を使用していない場合、「非造影」と訳す
- 「CTO」は「Cardiothoracic outline」のことで心臓のサイズや構造に言及している
- 「maximal physiological limit」は「生理的上限」と訳す
- 「natural」は「正常」と訳す
- 「lung parenchyma window」は「肺野条件」と訳す
- 「consolidation」はそのまま「コンソリデーション」と訳す
```

**System prompt (English translation)**

```
You are an expert translator specializing in radiology reports.
Translate the following English radiology report into natural, clinically accurate Japanese.

Constraints and notes:
- Do not alter the medical meaning
- Use standard Japanese radiology terminology
- Do not use colons or semicolons; use "。" and "、" instead
- Do not use the expression "断層内" (within the tomographic slice); use "撮像範囲内" or "検査範囲内" (within the imaging range) instead
- If no contrast agent was used, translate as "非造影" (non-contrast)
- "CTO" refers to "Cardiothoracic outline" and pertains to cardiac size and structure
- Translate "maximal physiological limit" as "生理的上限"
- Translate "natural" as "正常" (normal)
- Translate "lung parenchyma window" as "肺野条件"
- Translate "consolidation" as "コンソリデーション" (untranslated loan word)
```

**LLM Judge Prompt**
All three LLM judges (DeepSeek-V3.2, Mistral Large 3, and GPT-5) used the following system prompt and user prompt template at temperature 0, with JSON response format enforced. For each report pair, the two Japanese translations were assigned to positions A and B in randomized order; the assignment was recorded to allow post hoc normalization.

**System prompt (Japanese original)**

あなたは放射線科読影レポート翻訳の評価者です。
与えられた英語原文と 2 つの日本語訳（A/B）を比較し、どちらが優れているか、または引き分けかを判定してください。

重要:
- 英語原文への忠実性と医療安全性を最重視する
- Findings と Impressions は結合された文章として扱う

出力は厳密な JSON のみ。

**System prompt (English translation)**

You are an evaluator of radiology report translations.
Compare the given English source text and two Japanese translations (A/B), and judge which is superior or whether they are equivalent.

Important:
- Prioritize fidelity to the English source and medical safety above all
- Treat Findings and Impressions as a combined text

Output strictly in JSON format only.

**User prompt template (Japanese original)**

以下を評価してください。

# English (source)
{en_text}

# Japanese A
{ja_a}

# Japanese B
{ja_b}

評価項目:
1) 医療用語の正確さ
2) 日本語としての読みやすさ
3) 臨床でそのまま使用するとしたら、どちらがより適切か（総合）
4) 文体や表現の点で、どちらがより放射線科医が書いたレポートらしいか

出力要件:
- 1)〜4) の winner をすべて "A" / "B" / "TIE" から選ぶ

- 各項目に短い理由（日本語、1 文程度）を付ける
- 出力は以下の JSON 形式（キー名を厳守）

JSON 形式:

{
 "term_accuracy": {"winner": "A|B|TIE", "reason": "短い理由"},
 "readability": {"winner": "A|B|TIE", "reason": "短い理由"},
 "overall": {"winner": "A|B|TIE", "reason": "短い理由"},
 "radiologist_like": {"winner": "A|B|TIE", "reason": "短い理由"}
}

**User prompt template (English translation)**

Please evaluate the following.

# English (source)
{en_text}

# Japanese A
{ja_a}

# Japanese B
{ja_b}

Evaluation criteria:
1) Accuracy of medical terminology
2) Readability and fluency in Japanese
3) Overall clinical suitability (which would be more appropriate for direct clinical use)
4) Radiologist-style authenticity (which more closely resembles a report written by a radiologist in terms of style and expression)

Output requirements:
- Select the winner for criteria 1)–4) from "A", "B", or "TIE"
- Provide a brief reason for each criterion (approximately one sentence in Japanese)
- Output strictly in the following JSON format (key names must match exactly)

JSON format:

{
 "term_accuracy": {"winner": "A|B|TIE", "reason": "brief reason"},
 "readability": {"winner": "A|B|TIE", "reason": "brief reason"},
 "overall": {"winner": "A|B|TIE", "reason": "brief reason"},
 "radiologist_like": {"winner": "A|B|TIE", "reason": "brief reason"}
}